# AN AUTOMATED TAPE LAYING SYSTEM EMPLOYING A UNIAXIAL FORCE CONTROL DEVICE

Bernhard Rameder[1], Hubert Gattringer[1], Ronald Naderer[2] and Andreas Müller[1]

[1]Institute of Robotics, Johannes Kepler University Linz, Altenberger Straße 69, 4040 Linz, Austria
Email: {bernhard.rameder, hubert.gattringer, a.mueller}@jku.at,
Web Page: https://www.jku.at/en/institute-of-robotics/
[2]FerRobotics Compliant Robot Technology GmbH, Altenberger Straße 66c, 4040 Linz, Austria
Email: ronald.naderer@ferrobotics.at,
Web Page: https://www.ferrobotics.com/en/home/

**Abstract**
This paper deals with the design of a cost effective automated tape laying system (ATL system) with integrated uniaxial force control to ensure the necessary compaction forces as well as with an accurate temperature control to guarantee the used tape being melted appropriate. It is crucial to control the substrate and the oncoming tape onto a specific temperature level to ensure an optimal consolidation between the different layers of the product. Therefore, it takes several process steps from the spooled tape on the coil until it is finally tacked onto the desired mold. The different modules are divided into the tape storage spool, a tape-guiding roller, a tape processing unit, a heating zone and the consolidation unit. Moreover, a special robot control concept for testing the ATL system is presented. In contrast to many other systems, with this approach, the tape laying device is spatially fixed and the shape is moved accordingly by the robot, which allows for handling of rather compact and complex shapes. The functionality of the subsystems and the taping process itself was finally approved in experimental results using a carbon fiber reinforced HDPE tape.

## 1. Introduction

The use of UD tapes (unidirectional fiber reinforced tapes) enables the creation of lightweight components that consider load paths and guarantees an efficient use of the raw material. Certain reinforcing fibers like carbon or glass are embedded in a thermoplastic substrate, called matrix. Components manufactured using the so-called tape-laying process are becoming increasingly important in other areas such as prosthetics and the defense sector, in addition to the aviation and automotive industries. When applying the individual tape strips, it is important to precisely adhere to the melting temperatures and contact pressures for the respective thermoplastic material at all working points in order to obtain a good result. Therefore, regulating the temperature at the contact point and the contact force itself are of central importance [7]. The available heating systems are often relatively complex and therefore expensive [6, 7], and cost-intensive force sensors are often necessary for force control. Due to these circumstances, this paper deals with a concept and a test setup of an ATL system that uses cost-effective alternatives to conventional components normally used in the taping process and therefore reduces the overall costs. The quality of the end product not only depends on a well-designed ATL system, but also on an appropriate path planning strategy. Therefore, an additional chapter introduces a special control concept for a robot with opportunity to develop novel path planning strategies. Unlike many other systems, in this approach, the taper is spatially mounted and the shape, which holds the already tacked substrate, is pulled over it using an industrial robot. While taping, the tape gets unspooled due to the robot´s handling motions. This allows for handling of rather compact and complex shapes in contrast to most other tape laying systems.

## 2. Automated Tape Laying System Design

The presented system is divided into modules, which take over different tasks in the taping process. The main parts are at first a tape storage spool, which holds the raw material, followed by the tape processing unit, which is responsible for tape feeding, the guidance of the tape to the consolidation point and the tape cutting. Subsequently to these modules follows the heating zone and the force controlled consolidation unit, which finally tack the tape on the desired shape. Following sections give are more detailed overview on the named modules, which can be seen in Figure 1.

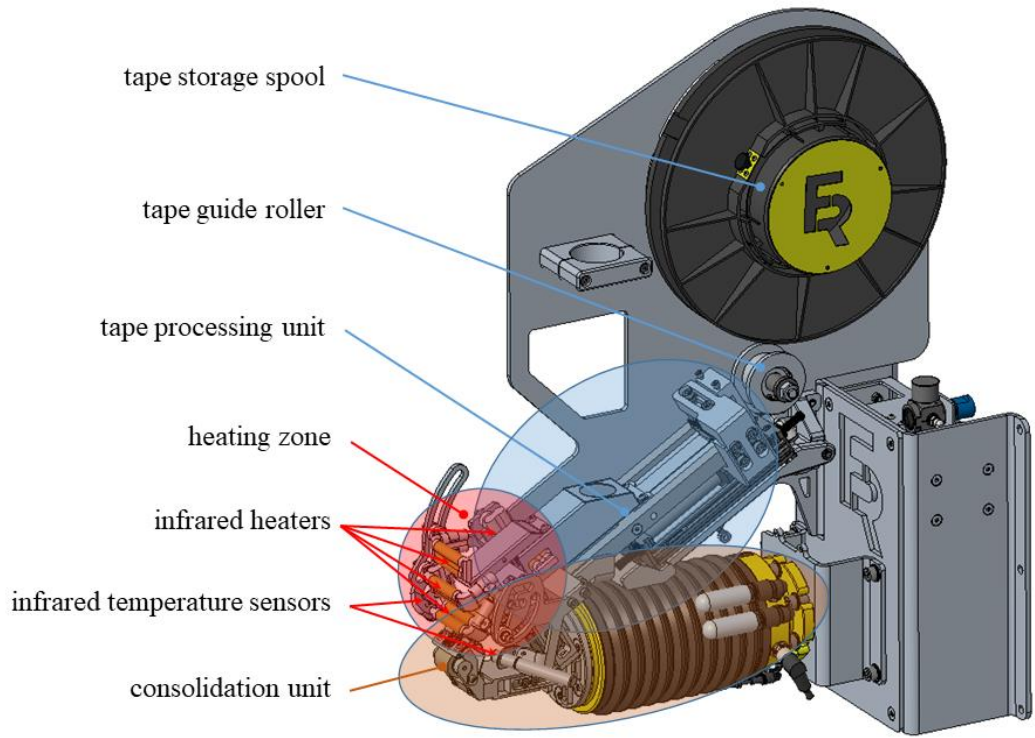

**Figure 1.** Module description of the automated tape laying system

### 2.1. Tape Storage Spool

The tape storage spool is designed to hold a cardboard ring, which takes up the raw UD tape stripe in its whole length. This tape holder can handle tape widths in a range between 0 mm and 51 mm. Spring-loaded elements on the back of the spool prevents that the tape gets unrolled after a motion. The outer ring can be fixed with a screw to keep the tape in position. An additional roller guide before the next unit guarantees a defined inlet of the raw material, independent from the current spool diameter.

### 2.2. Tape Processing Unit

A central module of the ATL system is the tape processing unit. It is responsible for tape feeding, the guiding and the cutting of the UD tape. Before the end of each laid track, the tape has to be cut a certain length before the actual track ends. This is realized with a toothed blade, which is actuated pneumatically via a set of miniature bellows cylinders. A plain bearing forces the blade to stay on its path to a groove. The retraction of the blade is done by spring elements. After a cutting process, the tape has to be fed forward to the consolidation roller again. This is done with a rodless pneumatic cylinder, which is magnetically coupled to a mechanically guided piston inside an air rail. The cylinder transfers its motion to the tape using a C-frame and a sleeve freewheel. The sleeve freewheel only allows a rotation in one direction, which feeds the tape in forward direction and allows the cylinder a back motion without pushing the tape back again. The end positions of the cylinder are monitored via auto switches and can be used to observe process conditions in the PLC program. Additionally, a second sleeve freewheel is placed near the tape inlet to prevent a remaining back motion of the tape caused by friction while the feed cylinder is moving to its rear position. A milled groove and several sheet metal parts guide the tape through the processing unit.

The tape processing unit also includes a preheating section, where the tape is brought to a temperature close beneath the processing temperature. This allows a more dynamic temperature control for the main heat zone in the nip point, the point where the tape is finally tacked on the mold. Figure 2 shows the part description of the tape processing unit.

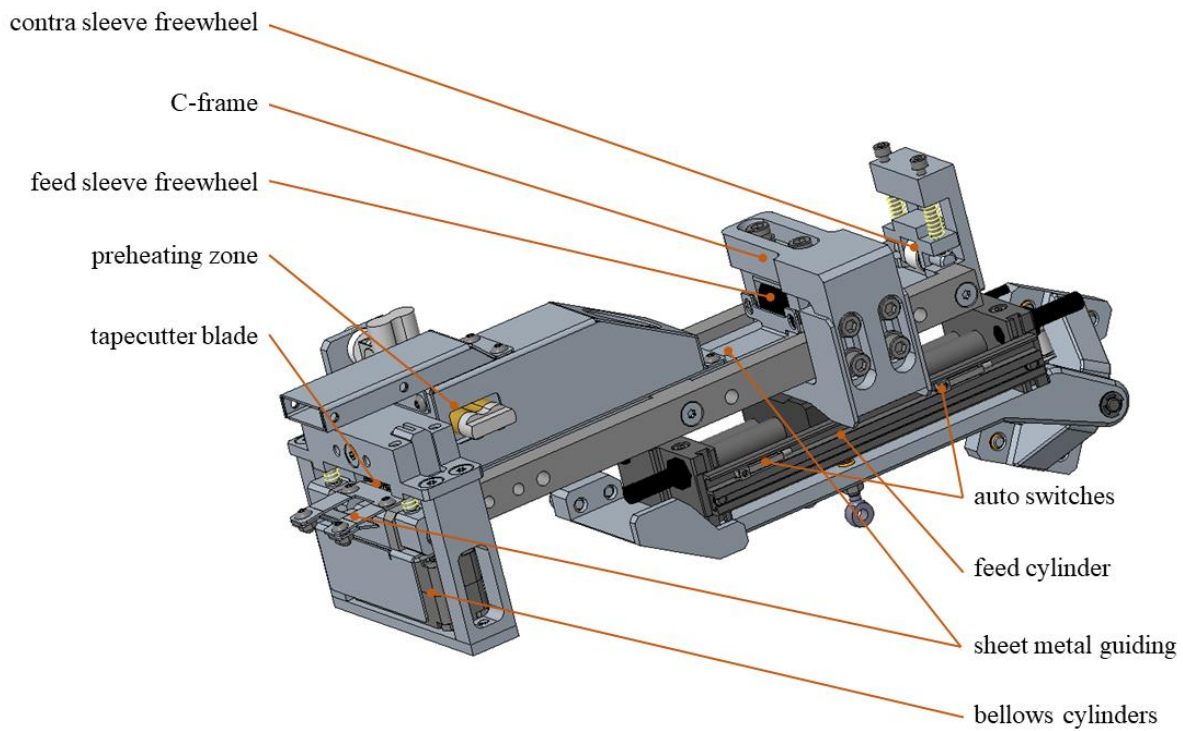

**Figure 2.** Part description of the tape processing unit

### 2.3. Heating Zone

One of the most critical parts by producing thermoplastic composite materials with the ATL process is the temperature control of the raw tape on the one hand and the already existing tape tracks on the mold on the other hand. To solve this problem the system has two separate temperature control units beside the preheating zone. Each of them consists of an infrared temperature sensor (Optris CS LT) and two infrared heaters with gold reflectors from Heraeus Noblelight, which are marked in Figure 1. A PID-controller is used to control the temperature measured by the infrared temperature sensor. The temperature controller provides an analog output signal for a current controller, which delivers the power for the infrared heaters. One of these heating systems is then used to control the temperature of the UD tape coming from the spool and the other one controls the heat on the already tacked tape stripes on the mold. For a satisfying compliance of tape and mold layers it is very important that both have the desired heat level. To this reason the temperature measurement takes place as close as possible before the nip point to ensure good results.

### 2.4. Consolidation Unit

Another crucial aspect of getting a good quality in the produced parts is to generate the desired contact force, which should ideally remain constant all over the current taping track. Therefore, in this ATL design a uniaxial force controlled system called ACF (active contact flange) [3] from FerRobotics is used. Several parameters as the desired force, payload or contact ramp and monitoring signals like actual stroke, contact state and error states can be set or read via Modbus. So once a desired contact force is set, it is controlled highly dynamic as long as you stay in the allowed area of the ACF's stroke, otherwise an error is thrown. A mechanical construction is mounted at the end of the ACF, which holds the so-called compaction or consolidation roller, respectively. This roller presses the heated tape on the surface of the mold and thereby it is tacked permanently on the desired shape. The consolidation roller has an additional layer of an elastic material on its outer diameter, which allows adapting to bumps on the mold's surface. Additional sheet metal parts guide the tape coming from the processing unit centered to the consolidation roller. Due to the integrated ACF, this concept has the great advantage that no force control needs to be implemented in the robot and an additional sensor can be saved. The construction is shown in Figure 3. With this approach, the system was originally designed to fix the tape on the desired surface in advance and then carry out the final processing, for example in an autoclave. However, the system has the ability to solidify the tape in situ, meaning no additional step is required after bonding to a specific substrate. The benefits of each method are stated in [4]. The influence of different process parameters and a comparison between in situ and autoclave products are covered in [9].

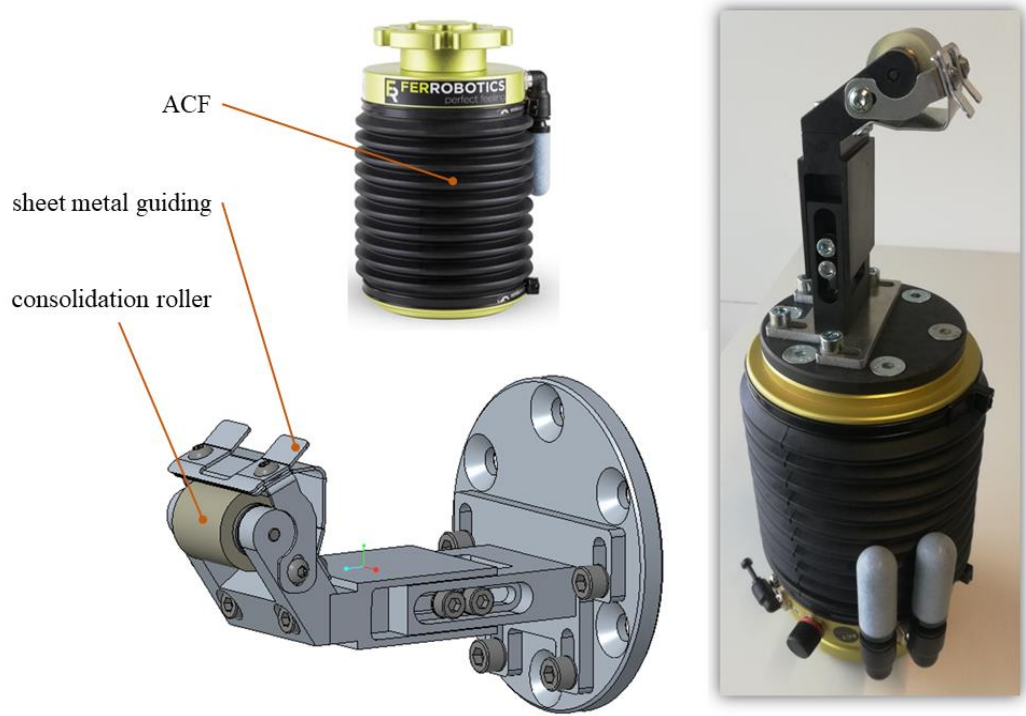

**Figure 3.** Part description of the consolidation unit

## 3. Test Setup
For testing the ATL system, a configuration with an ABB IRB 1600 robot and the taper was built up. Here, the mold on which the tape should be tacked is mounted on the end effector of the robot. The ATL system thereby is attached on a heavy plate beside the robot. This system is especially designed for producing small and complex composite parts. Therefore, it is easier to carry the mold instead of the relatively bulky taping device in comparison. As the robot in this case is not operated with the standard software from ABB, the robot control is handled in the following section separately. The experimental configuration can be seen in Figure 4.

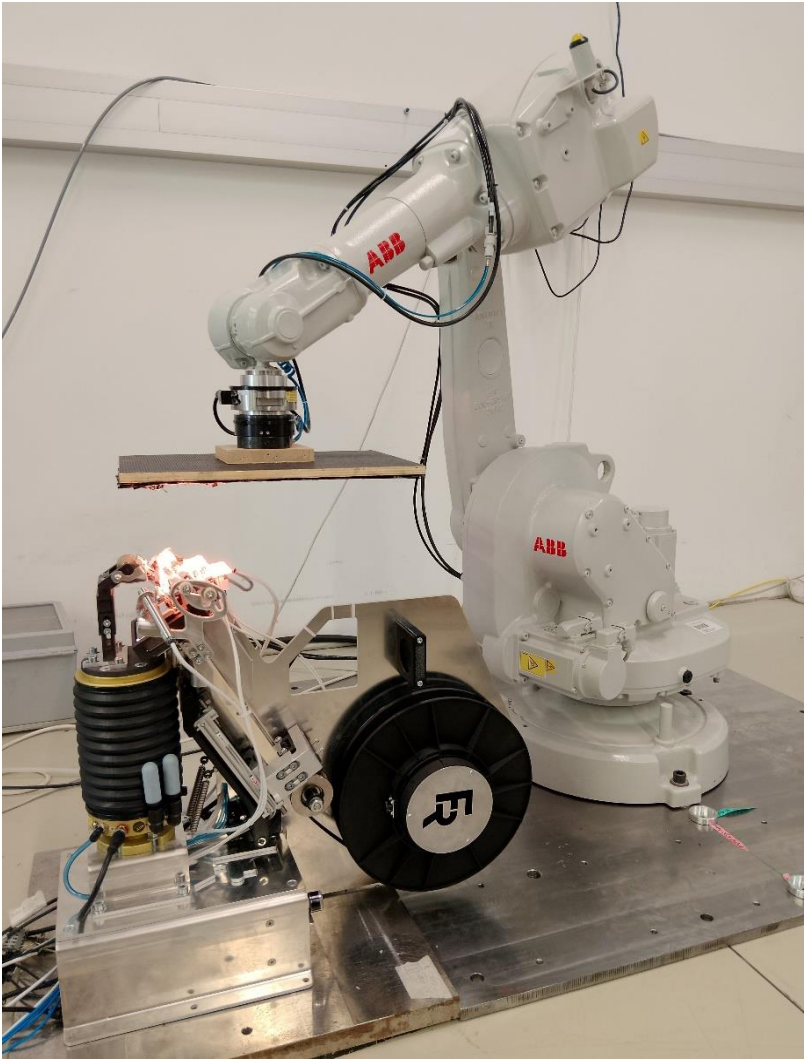

**Figure 4.** Test configuration with ATL system and handling robot

### 3.1. Robot Control ABB IRB 1600
For this approach, a robot from the type ABB IRB 1600 was used. However, in this case the robot was not operated with the standard electronics and operating program from ABB (RobotStudio). Instead, the robot was controlled with a B&R APC 910 industrial computer [2] with corresponding frequency converters called ACOPOS P3 [1]. A control scheme is shown in Figure 5. All substantial functions were then programmed in Automation Studio, the development environment of B&R. The path planning and the inverse kinematics were programmed in a simulation software and code generated for Automation Studio afterwards. This leads to a total freedom in developing new control strategies for the taping process as the path planning and optimization is not restricted to functionalities of RobotStudio. New approaches considering complex 3D-shapes or temperature and pressure models of the used thermoplastic matrix can be taken into account with this concept.

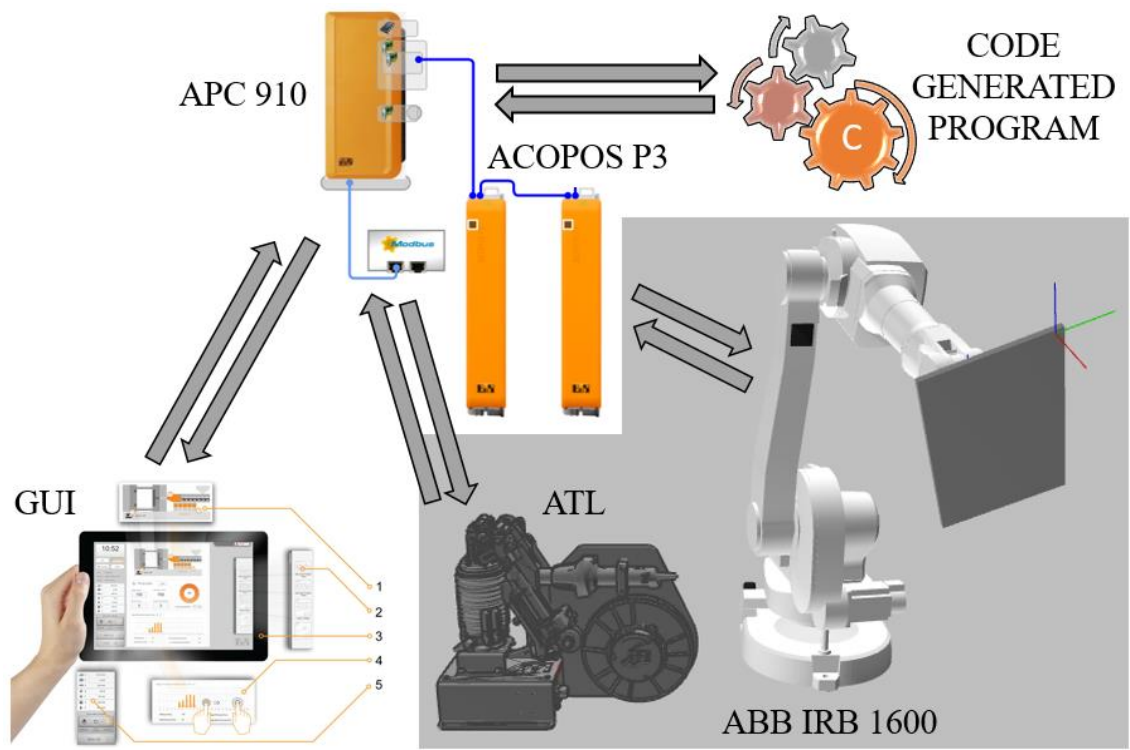

**Figure 5.** Robot and ATL system control scheme

### 3.2. ATL System Control
As well as the robot, the ATL system is also controlled using the APC 910 from B&R. Several digital in- and output modules are used to switch the valves for the installed pneumatics or the heating sources as analog modules manage the control of the temperature, as they convert input signals from the sensors to appropriate values for the PLC or write the desired control output to the corresponding controllers. As mentioned, the parameters for force control are transferred via Modbus.

### 4. Results
First experiments show promising results as can be seen in Figure 6. The tape stripes are tacked together all over the taping route. All components together ensure that the process runs automatically, once the correct system parameters like desired temperature and contact force are tuned accordingly. The tape stripes are melted together all over its surface and the developed robot control positioned them accurately. The first tests were done by using a plain wooden plate with an initial layer of the UD stripes on it. Process parameters were determined empirically for the used carbon reinforced HDPE tape.

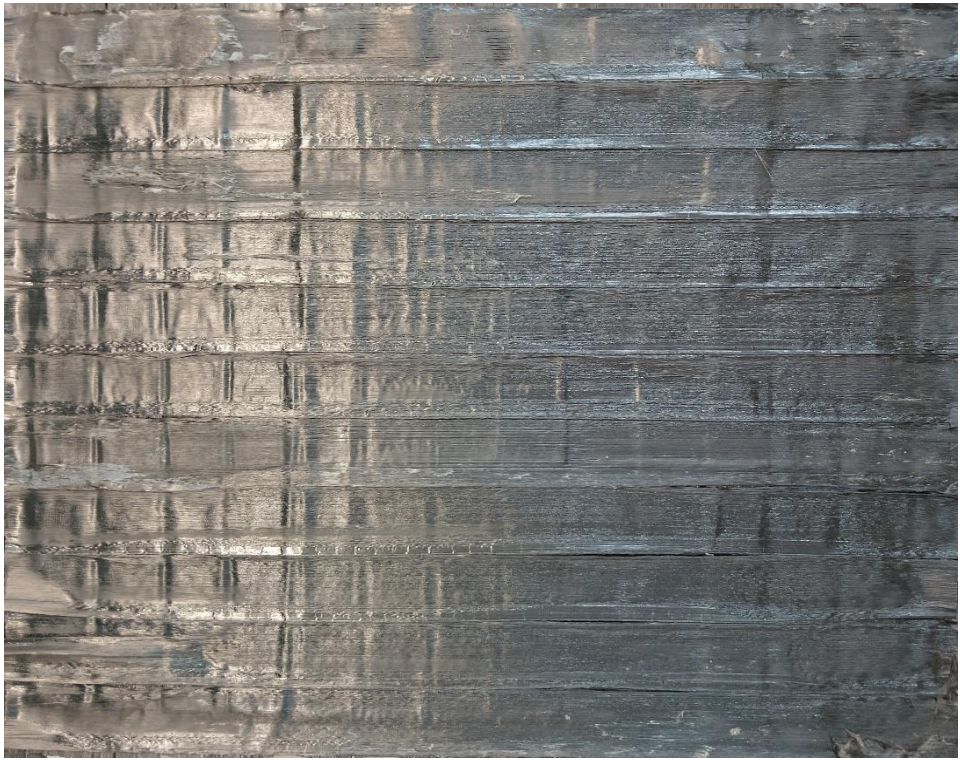

**Figure 6.** Result of a taping process on a plain plate with an initial UD tape layer

**5. Conclusion**
In this approach, good results were achieved in the experiments using the setup with ATL system and the self-made robot control. The tape stripes are fully tacked together and were positioned precisely. Compared to the components, which are normally used in such ATL systems, the installed parts are very cost effective and give the system a price advantage in comparison to commercial products. Furthermore, the in-house developed robot control facilitates wide opportunities to design novel control strategies, which enables improvements in the whole taping process. Therefore, some future investigations will be the consideration of the behavior of temperature and contact force related to the used thermoplastic matrix, which carries the carbon fibers of the UD tapes [7, 11]. Additionally, the heating power of the infrared emitters can be taken into account for calculating optimal trajectories for the taping process [8, 11]. Another approach would be a reinforcement learning for considering the process model, as it is proposed in [10]. In addition, the complexity of the mold's shape can be increased and new ways to handle the trajectory planning can be searched for this case as they suggest in [5].

**Acknowledgments**
This work has been supported by the "LCM - K2 Center for Symbiotic Mechatronics" within the framework of the Austrian COMET-K2 program.